\pdfoutput=1

\documentclass[11pt]{article}
\usepackage{amsmath}
\usepackage{acl}

\usepackage{times}
\usepackage{latexsym}
\usepackage{amsthm}
\usepackage[T1]{fontenc}

\usepackage[utf8]{inputenc}
\usepackage{algorithm2e}
\usepackage{microtype}
\usepackage{tabularx}
\usepackage{booktabs}
\usepackage{todonotes} 
\usepackage{enumitem}
\setenumerate[1]{label=[\arabic*]}

\title{GRAM: Fast Fine-tuning of Pre-trained Language Models \\
for Content-based Collaborative Filtering}

\author{Yoonseok Yang$^*$, Kyu Seok Kim$^*$, Minsam Kim$^*$, Juneyoung Park$^{\dagger{}}$ \\
        Riiid AI Research \\
        \texttt{\{yoonseok.yang, kyuseok.kim, minsam.kim, juneyoung.park\}} \\
        \texttt{@riiid.co}}

\begin{document}
\maketitle
\begin{abstract}

Content-based collaborative filtering (CCF) predicts user-item interactions based on both users’ interaction history and items’ content information. Recently, pre-trained language models (PLM) have been used to extract high-quality item encodings for CCF. However, it is resource-intensive to train a PLM-based CCF model in an end-to-end (E2E) manner, since optimization involves back-propagating through every content encoding within a given user interaction sequence. To tackle this issue, we propose \textbf{GRAM} (\textbf{GR}adient \textbf{A}ccumulation for \textbf{M}ulti-modality in CCF), which exploits the fact that a given item often appears multiple times within a batch of interaction histories. Specifically, \textbf{Single-step GRAM} aggregates each item encoding's gradients for back-propagation, with theoretic equivalence to the standard E2E training. As an extension of Single-step GRAM, we propose \textbf{Multi-step GRAM}, which increases the gradient update latency, achieving a further speedup with drastically less GPU memory. GRAM significantly improves training efficiency (up to 146$\times$) on five datasets from two task domains of Knowledge Tracing and News Recommendation. Our code is available at \href{https://github.com/yoonseok312/GRAM}{https://github.com/yoonseok312/GRAM}.




\end{abstract}

\section{Introduction}

\makeatletter
\def\blfootnote{\gdef\@thefnmark{}\@footnotetext}
\makeatother

\def\@fnsymbol#1{\ensuremath{\ifcase#1\or *\or \dagger\or \ddagger\or
  \mathsection\or \mathparagraph\or \|\or **\or \dagger\dagger
  \or \ddagger\ddagger \else\@ctrerr\fi}}

\blfootnote{*These authors contributed equally.}
\blfootnote{$^\dagger$This author is the corresponding author.}


Collaborative filtering (CF) is a popular technique used for mining relationships between items and users. Recently, CF has been successfully applied to various tasks including Knowledge Tracing and Recommender Systems ~\cite{recsysamazon, recsysnetflix, recsyshybrid}. However, conventional CF only considers item-user interactions, and disregards any item-/user-specific information. This leads to the so-called cold-start problem \cite{cfcoldstart}, where the CF model fails to make predictions for unseen users/items, even when they resemble observed users/items.

To remedy this issue, Content-based Collaborative Filtering (CCF) incorporates items’ content information into the item encoding. This not only addresses the cold-start problem, but also leads to significant performance improvement ~\cite{nrms, nrmsplm, coevoulutionary}. Specifically, large pre-trained language models (PLM) \cite{bert, gpt} have shown great potential for extracting items’ content information. However, fine-tuning PLMs for CCF requires prohibitive computational resources in terms of (1) training time and (2) GPU memory footprint.


This issue arises due to CCF's multi-modal nature, where item representations are learned from both tabular user records and their textual information. As a given item appears multiple times within a batch of users' records, its textual encoding needs to be computed every time it appears within the batch. Moreover, the need to store intermediate activations for hundreds of millions of parameters in PLM requires high GPU memory footprint \cite{e2ealternative}. 
 
To that end, we propose \textbf{Single-step GRAM}, \textbf{GR}adient \textbf{A}ccumulation for \textbf{M}ulti-modality in CCF, which alternately trains the task specific module and content encoder module. Accumulating gradients for repeated items in a training step, Single-step GRAM provides 4 times faster training while being theoretically equivalent to standard E2E training. 

As a natural extension of Single-step GRAM, we also propose \textbf{Multi-step GRAM} which accumulates gradients across \textit{multiple training steps}. Multi-step GRAM (1) provides an even higher ratio of acceleration without significant performance loss and (2) consumes less than 40\% GPU memory of E2E. Results show that the computational speed can increase up to 146$\times$ (45$\times$ on avg.) with less than 1\% drop in AUC.



The contributions of our paper are as follows:
\begin{itemize}[leftmargin=5.5mm, topsep=-2.5pt, itemsep=-2.5pt]
   \item We present GRAM (Single-step \& Multi-step) which accelerate training of CCF by accumulating the gradients of redundant item encodings.
   \item We empirically show that GRAM, especially Multi-step GRAM, noticeably reduces GPU memory footprint compared to E2E training.
   \item We evaluate GRAM in a variety of settings on 5 real-world datasets in two task domains, News Recommendation and Knowledge Tracing.
\end{itemize}

\section{Related Works}
\subsection{Collaborative Filtering and Content-based Filtering}
\label{sec2:ccf}


Collaborative filtering (CF) \cite{cfsurvey} attempts to predict user-item interaction based on past history. CF alone disregards any user-specific or item-specific information, leading to the inability to extract useful features from user information or item content. This especially leads to low performance on cold-start users and items. Various content-based filtering methods \cite{cbfsurvey, hybridsurveyunify} have been proposed to mitigate these issues through labeled meta-data. It uses raw textual features of item, instead of requiring other users' data during a user's recommendation like CF. However, these approaches require manual labelling and lack the extensive usage of the content itself.

\subsection{Content-based Collaborative Filtering}
Content-based Collaborative Filtering (CCF) incorporates content into CF in order to unify the strengths of CF and content-based filtering. It consists of two major components: content-encoder (CE) and collaborative filter (CF) components connected in an end-to-end fashion. Major task domains where items' raw textual content may significantly aid CF include:


\textbf{News Recommendation (NR)}, a task of predicting whether a user will click an article among others, provided with the user's past interactions. 


\textbf{Knowledge Tracing (KT)}, a task of predicting whether a user will correctly responds to a question or not based on the user's past responses. 

In this section, we briefly review the widely-used approaches to utilize content information in CCF, with a focus on thse two domains. 


\subsubsection{Training Content-Encoder (CE) in CCF}
Existing works mainly train or fine-tune CE module in an E2E fashion to obtain a useful content representations for the given CF task. In NR, NRMS \cite{nrms} applied Glove \cite{glove} word representation and Multi-head Self-attention (MHSA) \cite{attention} to encode the article's text to the item representation. Similarly in KT, EERNN \cite{eernn} used BiLSTM \cite{bilstm} to process Word2Vec \cite{word2vec} representations of question text into question representation. The representation is then fed into another LSTM layer to make final prediction on the user response. 

While aforementioned methods directly use content-encoder’s output as item vectors, there has also been work to use CE’s output to regularize item vectors. Topical Attention Regularized Matrix Factorization (TARMF) \cite{coevoulutionary} uses Matrix Factorization as CF and attention-based GRU network as CE to incorporate review data in predicting user-item ratings. Alternatively training CF and CE module, it uses the CE output to regularize the item representations in CF. 

Most recently, researchers started to fine-tune large Pre-trained Language Models (PLMs) with generic language understanding as a CE module for better content representation. In NR, NRMS-PLM \cite{nrmsplm} fine-tunes BERT \cite{bert} in an E2E manner, achieving meaningful performance gain. 

\subsection{Efficient Fine-tuning of Large PLMs}
While PLMs show powerful performance as a content-encoder in CCF, fine-tuning PLMs is known to be inefficient \cite{adapters} as it includes updating billions of parameters. Although it is possible to use PLM's output as fixed features for downstream tasks, numerous studies \cite{bert, sbert} emphasize such feature-based approach cannot match the performance of E2E fine-tuning. Thus, researchers have considered fine-tuning a subset of the PLM architecture \cite{bert} and adding task-specific parameters \cite{adapters} to reduce cost and performance degradation.

However, the computational complexity deteriorates even more under multi-modal settings like CCF. In such cases, PLM is called numerous times for a single-user, adding a new dimension of computational load, making E2E training almost impossible. SpeedyFeed \cite{speedyfeed} was proposed to accelerate the fine-tuning of PLM for news recommendation through combination of several methods. However, they are mainly engineering heavy implementations with domain dependencies. For a general training scheme in CCF, we propose a novel method that can be applied in an orthogonal manner to aforementioned techniques like \cite{adapters} and \cite{speedyfeed}, while achieving remarkable speed boost in such multi-modal CCF settings.

\section{Preliminaries}
\label{sec3:prelim}

\begin{table}[h]
\small
\begin{tabularx}{\columnwidth}{l|l}
\toprule
Notation               & Description                                                                           \\ 
\midrule
$I^u$                  & Interaction sequence of user $u$                                                      \\
$c_i$                  & Content of item $i$                                                                   \\
$I^u_n=(e_n^u, r_n^u)$ & $n$-th interaction of user $u$                                                        \\
$e_n^u=CE(c_n^u)$      & Embedding of content $c_n^u$ from CE                                            \\
$B^{(t)}$                & Mini-batch $B$ at update time-step $t$                                                \\
$[B_I, B_u]$           & Set of unique [items, users] in $B$                                        \\
$[l_t, l_I]$           & Length of [tokens, interactions] in  [$c$, $I$] \\
$d$                    & Content embedding dimension  
\\
$\mathcal{L}$       & CCF minimization objective
\\
\bottomrule
\end{tabularx}
\normalsize
\caption{ \label{tab:notations}
Notations for Content-based CF(CCF)
}
\end{table}

In this section, we formally setup CCF framework and its notations for efficient discussion. CCF framework consists of two major components: content-encoder (CE) and collaborative filter (CF) components connected in an end-to-end fashion. 

\subsection{CF component} 
CF component predicts user $u$'s response $r$ to an arbitrary item based on the user's past interactions $I^u=(I^u_1,I^u_2,...,I^u_{|I^u|})$ where each $n$-th interaction $I^u_n=(e^u_n,r^u_n)$ is represented as a tuple of item representation $e^u_n$ and the user's response to the item $r^u_n$. In other words, the CF module aims to estimate the probability:

\vspace{-0.1in}
\small
$$CF(I^u;e^u_{|I^u|+1}) = {P}(r^u_{|I^u|+1} | I^u_1,I^u_2,...,I^u_{|I^u|};e^u_{|I^u|+1})$$
\normalsize


\subsection{CE component} 
CE component outputs the item representation $e$: $e^u_n = CE(c^u_n)$ where $c^u_n$ is the token sequence of the corresponding item. The model parameters $\theta_{CE}$ and $\theta_{CF}$ of the CE-CF pipeline is then trained in an end-to-end fashion based on cross-entropy loss for response prediction.
The summary of notation used for CCF is provided in Table \ref{tab:notations}. The existing approaches to tackle CCF are formatted and presented in Table \ref{tab:modelchoice}, along with the pipeline we adopt for our later experiments.

\begin{table}[h]
\small
\begin{tabularx}{\columnwidth}{c|c|cc}
\toprule
Task                & Model           & CE                    & CF         
\\ \hline
 & NRMS            & Glove, MHSA & MHSA \\
 NR                   & NRMS-PLM        & BERT                  & MHSA \\
                    & Our Experiments & BERT                  & MHSA \\ 
\midrule                    
  KT                  & EERNN           & W2V, BiLSTM           & LSTM           \\
                    & Our Experiments & BERT                  & LSTM \\
\bottomrule
\end{tabularx}
\normalsize
\caption{ \label{tab:modelchoice}
CE-CF Pipeline Choice for CCF
}
\end{table}

\subsection{Inefficiency of E2E in CCF}
In CCF, end-to-end fine-tuning of the CE (PLM) suffers from cubic computational complexity in terms of sequence length, due to the data multi-modality. Let's assume average text token length of $l_t$ and average interaction record length of $l_I$ with each mini-batch $B$ containing users $B_u$.

Attention-based CE module would be called $l_I$ times, producing forward/backward-pass computational complexity of $O(|B_u| \cdot l_I(l_t^2d + l_td^2))$ where $d$ represents embedding dimension. 

Under similar average sequence length of per-item tokens and per-user interactions $l_t \approx l_I$, the resulting cubic complexity in terms of sequence length significantly increases space and time complexity of model training and becomes the limiting bottleneck factor.



\section{Proposed Method}
\label{sec4}

For efficient training, we propose \textbf{GR}adient \textbf{A}ccumulation for \textbf{M}ulti-modality in CCF (GRAM) with two variants: \textbf{Single-step GRAM} and \textbf{Multi-step GRAM}. 

\begin{figure*}[t]
    \centering
    \includegraphics[width=1.7\columnwidth]{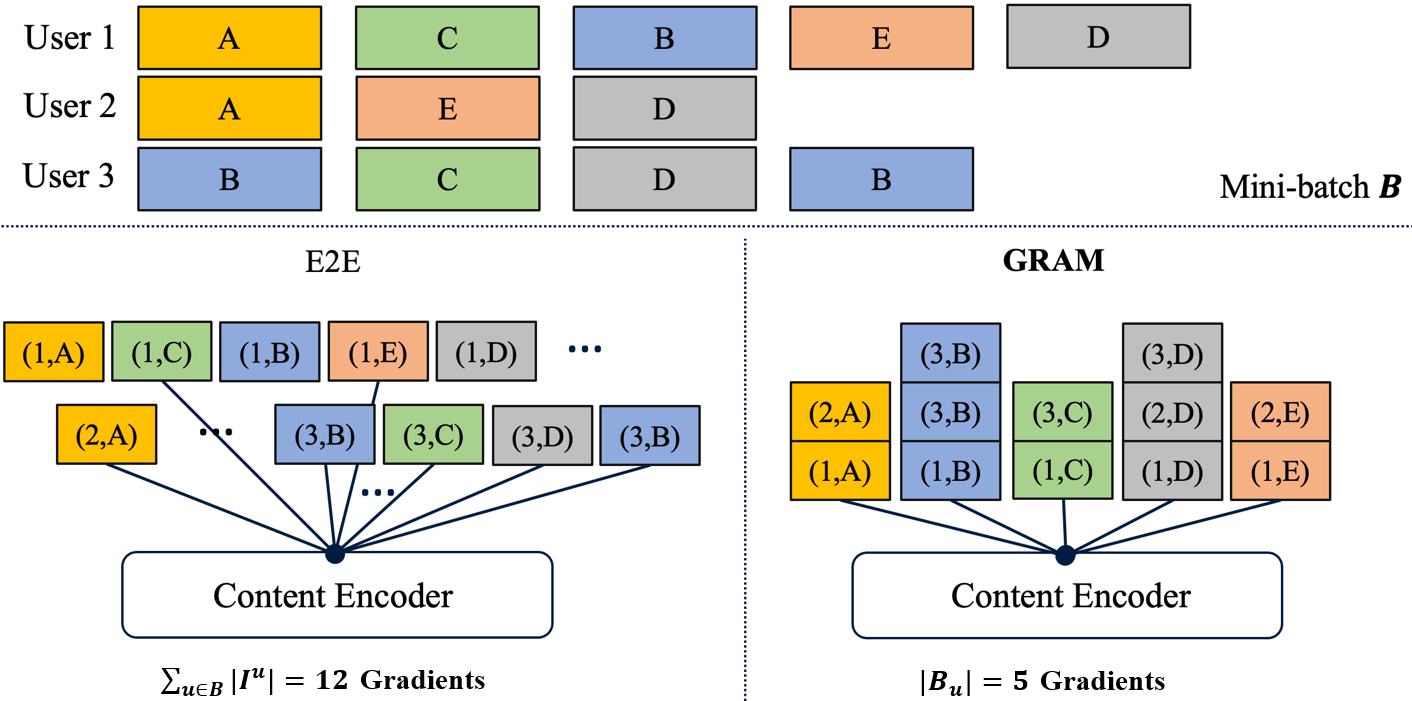}
    \caption{Comparison to E2E Training: Let's assume a mini-batch $B$ of 12 interactions consisting of 3 users and 5 unique items. For the content encoder, E2E computes 12 gradients from each \textit{interaction}, while GRAM computes 5 only, based on accumulated gradient signal in pseudo-target $\tilde{h_i}$ from each \textit{item} $i$.}
    \label{fig:backprop}
\end{figure*}

\subsection{Single-step GRAM}
\label{sec4:gram}
Single-step GRAM trains CE module and CF module \textit{separately}. To update the CE module, Single-step GRAM accumulates gradients of redundant items, effectively reducing the number of gradient computation for each step. As shown in Figure \ref{fig:backprop}, Single-step GRAM can accelerate training by limiting the number of gradient calculation to the number of unique items in the batch. With this, Single-step GRAM can reduce the algorithmic complexity of CE to $O(|B_I| \cdot (l_td^2 + l_t^2d))$ per mini-batch update, from E2E's $O(|B_u| \cdot l_I(l_t^2d + l_td^2))$. 

\newtheorem*{theorem}{Proposition}
\begin{theorem}
Given parameters $(\theta_{f},\theta_{g})$ , suppose a neural network in separable form with $y=g(h;\theta_g), \space h = f(x; \theta_f)$, and loss $L(\theta_f, \theta_g)$. Define 
pseudo-target $\tilde{h}$ as
\begin{align}
\tilde{h} &:= 
f(x;\theta_f) - \frac{\partial L}{\partial h}\bigg|_{h=f(x;\theta_f)},
\label{eqthm:pseudotarget}
\end{align}
\\
and pseudo-loss $\tilde{L}(\theta_f)$ for another network $f(x; \theta_f')$ as
\begin{align}
\tilde{L}(\theta_f) &:= 
\frac{1}{2} 
\bigg(
\tilde{h} - f(x;\theta_f')
\bigg)^2.
\label{eqthm:pseudoloss}
\end{align}

Then, the gradient of $\theta_f$ w.r.t $L$ is equivalent to the gradient of $\theta_f'$ w.r.t $\tilde{L}$,

\begin{equation}
    i.e. \quad \frac{\partial L}{\partial \theta_{f}} = \frac{\partial \tilde{L}}{\partial \theta_f'}\bigg|_{\theta_f'=\theta_f}.
    \label{eqthm:equivalence}
\end{equation}
\end{theorem}

While GRAM trains CF module and CE module separately, it guarantees a theoretically equivalent parameter update with E2E as shown in the Proposition (proof in Appendix \ref{appendix:proof}). Yielding identical outcome of E2E back-propagation with fewer gradient computation, Single-step GRAM enables efficient training under hierarchical multi-modal setting like CCF by accumulating gradient buffer in the pseudo-target for items that are updated multiple times in a mini-batch. 

\subsection{Multi-step GRAM} 
Instead of a single update, we can accumulate gradient buffers for \textit{multiple} mini-batch updates, gaining additional speed boost. In other words, CF / CE modules may separately update multiple times across multiple mini-batches, further elaborated in the following section. However, this relaxation of GRAM's alternating period no longer guarantees equivalence with E2E back-propagation.

As the CE module can use multiple steps to update item representations, Multi-step GRAM can also reduce GPU memory footprint drastically compared to E2E. 


\subsection{Training Scheme of GRAM}
\label{sec4:gramalgo}

\RestyleAlgo{ruled}

\SetKwComment{Comment}{/* }{ */}
\DontPrintSemicolon 
\begin{algorithm}[hbt!]
\SetKwInOut{Input}{Input}
\SetKwInOut{Output}{Output}
\caption{GRAM}\label{alg:two}
\Input{$\{B_I^{(t)}\}_{t=0}^{T-1}$: mini-batch,\\ $\theta^{(0)}$: initial parameters}
\Output{$\theta^{(T)}$: resulting parameters}

\;
\While{$t = \{0,...,T-1\}$}{

    (1. Produce content representations)
    \begin{equation}
    h^{(t)}_i \gets CE(c_i;\theta_{CE}^{(t)}), \forall i \in B_I^{(t)}
    \label{eq:produceitem}
    \end{equation}
    (2. Update CF parameters)
    \begin{equation}
        \theta_{CF}^{(t+1)} \gets \theta_{CF}^{(t)} - opt(\nabla \mathcal{L}(\theta_{CF}^{(t)})) 
        \label{eq:updatecf}
    \end{equation}
    (3. Update content representations)
    \begin{equation}
        \tilde{h}_i^{(t)} \gets h_i^{(t)} - \nabla \mathcal{L}(h_i^{(t)}), \forall i \in B_I^{(t)}
        \label{eq:updateitem}
    \end{equation}
    \If{$t \mod N = 0$}
    {
    (4. Update CE parameters)
    \begin{equation}
        \theta_{CE}^{(t+1)} \gets \theta_{CE}^{(t)} - opt(\nabla \tilde{\mathcal{L}}(\theta_{CE}^{(t)}))
        \label{eq:updatece}
    \end{equation}
    }
    
    $t \gets t+1$
}
\end{algorithm}

 See Algorithm 1 for the pseudo-code of GRAM. We denote gradient accumulation step as $N$. The algorithm becomes Single-step GRAM for $N = 1$ and it becomes Multi-step GRAM for $N > 1$. 
 
 First, CE produces content representation $h_i^{(t)}$ for CF module to complete the forward-pass for unique items $i$ in the mini-batch $B^{(t)}$. Then, we update the CF module's parameters using standard back-propagation with a normal CF training objective $\mathcal{L}$ in Eq.\eqref{eq:cfloss}, while simultaneously updating the content representation (output from CE module) by treating it as trainable embedding $\tilde{h}_i^{(t)}$.

\begin{align}
\mathcal{L} &:= \mathlarger{\sum}_{u \in B_u^{(t)}}
        L(I^u, \{h^{(t)}_i|\forall i \in B_I^{(t)}\};\theta_{CF}^{(t)})
\label{eq:cfloss}
\end{align}

Lastly, for items of which embeddings are modified in Eq.\eqref{eq:updateitem}, the CE module is trained to follow/regress towards the modification with respect to pseudo-target $\tilde{h}_i^{(t)}$ and pseudo-loss $\tilde{L}$ in Eq.\eqref{eq:pseudoloss}.

\begin{equation}
    \tilde{\mathcal{L}} := \frac{1}{2}\mathlarger{\sum}_{i \in B_I^{(t)}}(\tilde{h}_i^{(t)} - CE(c_i;\theta_{CE}^{(t)}))^2
    \label{eq:pseudoloss}
\end{equation}

Based on updated content representations, we repeat the process with $t \leftarrow t+1$. Note that for Eq.\eqref{eq:updateitem}, Stochastic Gradient Descent with a learning rate of 1 should be used to guarantee the theoretical equivalence with E2E. For  Eq.\eqref{eq:updatecf} and Eq.\eqref{eq:updatece}, choice of a optimizer (e.g. Adam) doesn't impact the equivalence with E2E.

\subsection{GRAM's Speed Boost ratio $\mathcal{R}$} 

 Given a mini-batch $B$ of $|B_u|$ users' interaction sequences, the standard E2E back-propagation updates CE module for $\sum_{u \in B_u}|I^u|$ (i.e. number of total interactions), while Single-step GRAM updates CE module for $|B_I|$ (i.e. number of unique items) times only. Since PLM based CE modules are usually significantly larger than the head (CF module) attached for downstream task such as News Recommendation, the following ratio of speed boost for CE module applies in a close to directly proportionate manner for the entire training procedure.

\begin{equation}
    \mathcal{R}:= \frac{\sum_{u \in B_u}|I^u|}{|B_I|} = \frac{\textrm{\#interactions($B$)}}{\textrm{\#items($B$)}}
    \label{eq:boostratio}
\end{equation}

The ratio $\mathcal{R}$ monotonically increases as mini-batch size becomes larger. Thus, larger mini-batch size would yield larger efficiency boost via Single-step GRAM. This is why Multi-step GRAM can achieve even more speed boost compared to Single-step GRAM. If the gradient accumulation latency becomes 1 epoch for Multi-step GRAM, the speed boost ratio $\mathcal{R}$ becomes:

\begin{equation}
    \mathcal{R}:= \frac{\textrm{\#total interactions in dataset}}{\textrm{\# total items in dataset}}
    \label{eq:boostratio}
\end{equation}

Considering there are significantly less number of items compared to the total number of interactions in real-world datasets, GRAM with high enough update latency can achieve remarkable speed boost. However, we can also expect that longer accumulation latency would hurt model performance and convergence. In the following section's experiments, (i) training efficiency boost and (ii) performance degradation from different alternating frequency are closely monitored on various GRAM alternating periods (Single-Step, 10-Step, Half-Epoch, Full-Epoch).

While GRAM utilizes gradient accumulation across duplicate item representations to boost training, the resulting speed boost is orthogonal with traditional gradient accumulation as it focuses on increasing effective batch size under limited computational resource. 


\begin{table*}[ht]
\small
\centering
\begin{tabularx}{1.8\columnwidth}{c|cccccc}
\toprule
Dataset & Users   & Total Items & Total Interactions & CS Items & CS Interactions & Average $l_t$ \\
\midrule
Spanish & 2,643    & 4,628         & 279,747                    & 200              & 3,191                         & 5.32                              \\
French  & 1,202    & 4,078         & 174,749                    & 200              & 1,970                         & 5.24                              \\
POJ     & 22,110   & 2,597         & 898,384                   & 200              & 10,523                        & 271.34                   \\
TOEIC   & 1,240,955 & 9,336         & 94,264,845                 & 684              & 321,933                       & 147.47                            \\
MIND    & 750,434  &  104,150        & 3,760,125                  &                  &                              & 11.52(639.57) \\
\bottomrule
\end{tabularx}
\normalsize
\caption{\label{tab:datameta} Dataset Information. For MIND, both average $l_t$ of title alone and $l_t$ of title + abstract + body are reported.}
\end{table*}

\section{Experimental Settings}
We first define the scope of tasks and metrics used in the experiments. Detailed description of datasets and methods are provided in section \ref{sec4:datasets} and section \ref{sec4:methods}, respectively.

\noindent
\textbf{Tasks: }Experiments are conducted on two major task domains of CCF: Knowledge Tracing(KT) and News Recommendation(NR), where models predict whether a student/reader will solve/click an question/article, as a classification task.

\noindent
\textbf{Metrics: }Overall AUC and cold-start item AUC (CSAUC) are measured for KT. AUC, MRR, nDCG@5, nDCG@10 are measured for NR. As cold start problem is intrinsically abundant in news recommendation environment (\citet{nrms}), we did not measure CSAUC separately. Wall-clock training time until convergence is reported for all experiments. For fair comparison of training time, all models are run on equivalent device (NVIDIA A100 GPU) in an isolated manner. 

\subsection{Experimental Details}
\label{sec4:datasets}
To evaluate our model, we used five real-world datasets: four datasets in KT, and one dataset in NR, on which both textual content data and user interaction data are available.
Experiments on Duolingo French and Spanish dataset are done with single NVIDIA A100 GPU, and those on POJ, TOEIC, and MIND are done with eight NVIDIA A100 GPUs, in distributed data parallel training. The results were shown to be statistically significant (p < 0.05).

Detailed per-dataset description is written below, and specification is in Table \ref{tab:datameta}. \footnote{Train / Test set were randomly split in 8:2 ratio. 10\% of train dataset were randomly chosen for validation set.} Dataset collection methods are mentioned in Appendix \ref{appendix:dataset_collection}. 

\subsubsection{Datasets}

\noindent
\textbf{(KT) Duolingo (Spanish, French): Language Translation} \cite{duolingo} contains questions and responses for Duolingo users. Following \citep{lmkt}, we collapsed the original token level mistakes to question level binary labels. We used Spanish and French dataset. 

\noindent
\textbf{(KT) POJ: Computer Programming} was collected from Peking online platform and consists of computer programming questions.

\noindent
\textbf{(KT) TOEIC: Language Comprehension} from EdNet \citep{ednet} is the largest publicly available benchmark dataset in education domain consisting of student interaction logs. 

\noindent
\textbf{(NR) MIND: Microsoft News Dataset} is one of the largest English dataset for monolingual news recommendation. MIND dataset provides news articles' title, abstract, and body text for news content modeling. For comparison with other models, we only utilized the news title, following  \citet{nrms, nrmsplm}. As Multi-step GRAM's significant speed boost allowed us to utilize more features, we also provide results on utilizing all textual features for Multi-step GRAM as well. 

\subsection{Baseline Methods} 
\label{sec4:methods}

\subsubsection{E2E \& GRAM}
\label{sec4:basemodel}
To fairly compare GRAM's training methodology against the standard E2E, we apply GRAM/E2E on an identical model architecture, defined for KT and NR respectively. Model choices are shown in Table \ref{tab:modelchoice}. To the best of our knowledge, this work is the first study to fine-tune BERT for KT. Detailed model architectures are described in Appendix \ref{appendix:detail}.

Based on this model architecture, we compare: \textbf{E2E} training, single-step \textbf{GRAM 1S}, 10-Step \textbf{GRAM 10S}, 0.5-Epoch \textbf{GRAM 0.5E}, and 1-Epoch \textbf{GRAM 1E} on the aforementioned metrics. We also provide benchmarks for other existing approaches, as elaborated in the following section \ref{sec4:othermodels}. Note that E2E training in NR has the same structure as NRMS-PLM in \cite{nrmsplm}.

\subsubsection{Other Baselines}
\label{sec4:othermodels}
In addition to the shared CCF model architecture defined above, we also include other approaches for extensive comparison. 

\noindent
\textbf{NoFinetune} approach directly adopts fixed item representation encoded from PLM without fine-tuning. Only the CF component is trained, receiving PLM's fixed output as the input. 

\noindent
\textbf{NoContent} approach does not incorporate any textual content at all. Each item representation is randomly initialized before being trained along with the standalone CF component. For KT, we used DKT ~\cite{dkt}.


\noindent
\textbf{(KT) Content Regularized CF (CRCF)} is our implementation of the proposed regularization in TARMF with equivalent CF and CE modules of GRAM. As TARMF’s content-encoder and user feature vector should go through additional iterations of optimization in sequential recommendation, we adopt hierarchical user encoder like E2E setting to eliminate the need of model retraining. 

\noindent
\textbf{(KT) LM-KT} formulates KT as auto-regressive modeling task to fine-tune pre-trained GPT-2. The method's major bottleneck to other sequential recommendation domains is that the model's sequence length has to increase in a multiplicative fashion: $O(l_t \times l_I)$ in Table \ref{tab:notations}. \footnote{LM-KT baseline was only tested on Duolingo datasets as other datasets' large average token length prevents LM-KT from considering more than a few items per each user.}

\noindent
\textbf{(KT) EERNN} is a specific instance of CE-CF pipeline where Bi-directional LSTM (CE) encodes question text's W2V representation into question embedding. CF consists of another LSTM layer. 

\noindent
\textbf{(NR) NRMS} is another specific instance of CE-CF pipeline where CE uses Glove word embedding and Multi-Head Self-Attention layers. Its CF component is also based on MHSA layers.

\noindent
\textbf{(NR) SpeedyFeed} used (i) auto-regressive modeling, (ii) BusLM, (iii) Dynamic Batching, and (iv) Cache mechanism to speedup PLM-based news recommendation. For fair comparison, we used equivalent CE module for both GRAM and SpeedyFeed.

    

Further experimental details are in Appendix \ref{appendix:detail}.


\begin{table}[h]
\centering
\small
\begin{tabular}{c|r|ccc}
\toprule
Dataset & Method     & AUC    & CSAUC & Speed-up \\ \midrule
TOEIC   & E2E        & 75.7   & 64.2  & 1(135hr) \\
        & GRAM 1S    & \textbf{76.0}     & 63.0    & 5.7  \\
        & GRAM 10S   & 75.8   & \textbf{65.1}  & 1.7    \\
        & GRAM 0.5E  & 75.7   & 64.7  & 26     \\
        & GRAM 1E    & 75.3   & 64.6  & \textbf{146}    \\
        & EERNN      & 75.8   & 62.3  & 10     \\
        & NoFinetune & 69.1   & 64.3  & 343    \\
        & NoContent  & 74.4   & 49.4  & 2547   \\ \midrule
POJ     & E2E        & 69.0   & 65.4  & 1(123m)  \\
        & GRAM 1S    & 69.0   & 65.5  & 4.5      \\
        & GRAM 10S   & \textbf{69.1}   & 65.0    & 3.8      \\
        & GRAM 0.5E  & 68.7   & 64.7  & 9.1      \\
        & GRAM 1E    & 68.8   & 64.5  & \textbf{12.5}      \\
        & EERNN      & 68.6   & 64.0    & 1.3      \\
        & NoFinetune & 68.3   & \textbf{65.8}  & 41.0     \\
        & NoContent  & 63.8   & 50.9  & 30.8     \\ \midrule
Spanish & E2E        & \textbf{75.1}   & \textbf{68.7}  & 1(74m)   \\
        & GRAM 1S    & \textbf{75.1}   & 68.3  & 3.5      \\
        & GRAM 10S   & 74.4   & 67.2  & 2.1      \\
        & GRAM 0.5E  & 74.4   & 67.0    & 2.8      \\
        & GRAM 1E    & 74.7   & 67.0    & \textbf{5.7}      \\
        & CRCF 1S    & 74.9   & 66.2  & 1.2      \\
        & CRCF 1E    & 74.3   & 66.3  & 2.1      \\
        & EERNN      & 74.3   & 66.3  & 1.6      \\
        & LM-KT      & 74.6   & 68.7  & 0.5      \\
        & NoFinetune & 72.5   & 66.2  & 24.7     \\
        & NoContent  & 67.0   & 49.3  & 37.0     \\ \midrule
French  & E2E        & 74.8   & 74.7  & 1(39m)   \\
        & GRAM 1S    & \textbf{75.0}   & \textbf{75}    & 3.3      \\
        & GRAM 10S   & 74.2   & 73.2  & 3.5      \\
        & GRAM 0.5E  & 74.1 & 73.2  & 4.3      \\
        & GRAM 1E    & 74.4 & 73.3  & \textbf{7.8}      \\
        & CRCF 1S    & 74.7 & 73.0    & 1.0      \\
        & CRCF 1E    & 74.4 & 72.9  & 1.8      \\
        & EERNN      & 74.0 & 71.3  & 1.2      \\
        & LM-KT      & 74.3 & 74.7  & 0.3      \\
        & NoFinetune & 71.4 & 69.4  & 9.8      \\
        & NoContent  & 67.0 & 49.5  & 13.0     \\ 
\bottomrule
\end{tabular}
\caption{\label{tab:kt} Prediction Performance / Speed-up in Knowledge Tracing. Training time is reported for E2E training, and best results among content finetuning methods are marked in bold.}
\normalsize
\end{table}

\begin{figure*}[h]
    \centering
    \includegraphics[width=1.7\columnwidth]{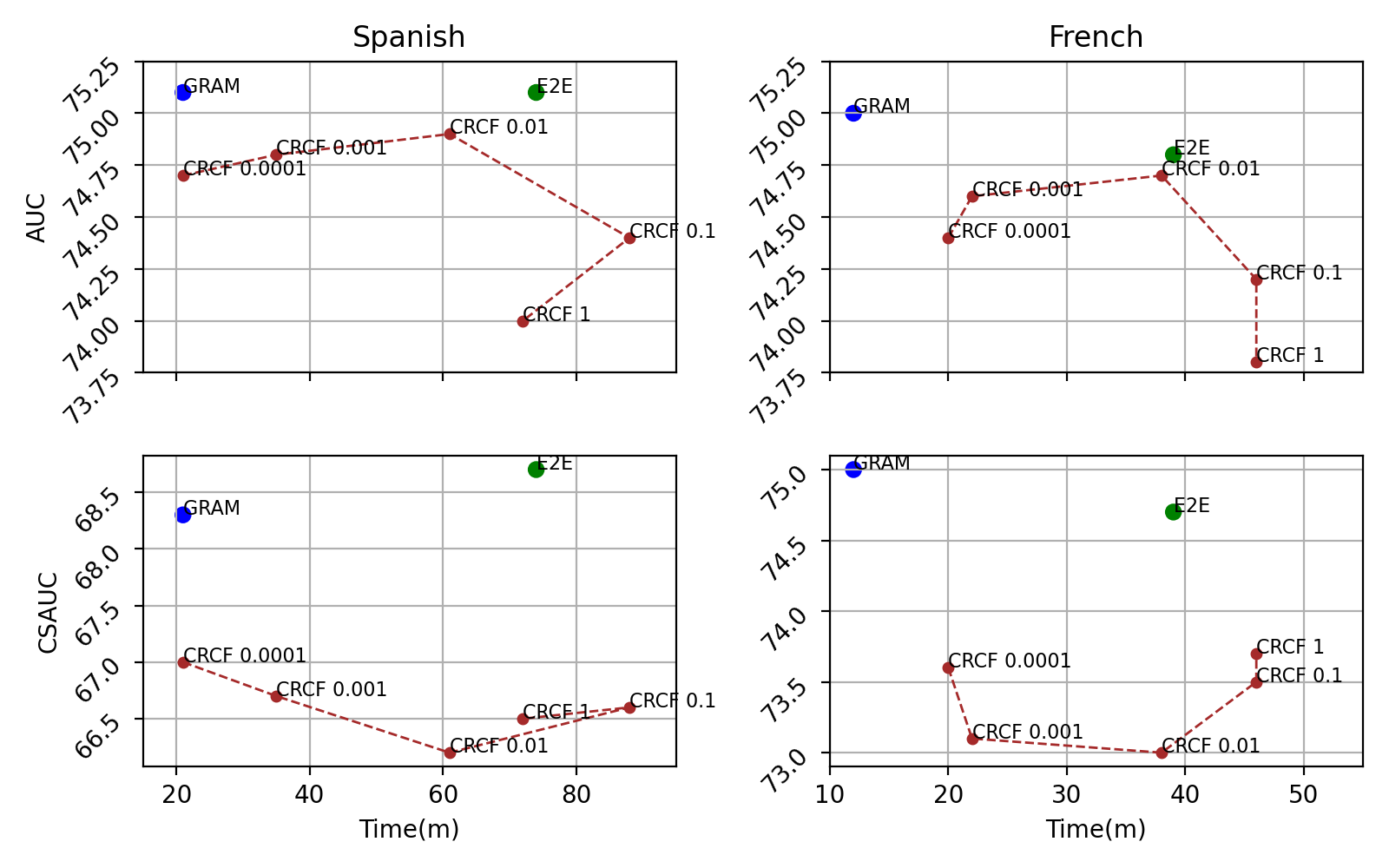}
    \caption{\label{fig:crcf} Comparison between GRAM, E2E, and CRCF. For CRCF, regularization parameter is marked as well.}
\end{figure*}

\section{Results and Discussion}

\subsection{Knowledge Tracing}

\noindent \textbf{Performance Comparison: } In general, E2E and GRAM in Table \ref{tab:kt}\footnote{For CRCF, best result is reported among hyper-parameter ablations in Figure \ref{fig:crcf}.} shows best performance across datasets. Difference among GRAM variants highlights how Multi-step GRAM's performance gradually deviates from the E2E baseline as alternating period increases. As claimed in the previous section \ref{sec4:gram}, GRAM(1S)'s performance most closely matches the E2E baseline. Confidence intervals of the two methods are reported in Appendix \ref{appendix:result_detail}. 

NoFinetune and NoContent achieve significantly worse performances on both metrics on all datasets, as compared to the full CCF setting with both CE and CF fine-tuned properly. Also, NoContent does not show any inference power on cold-start items, reporting AUC values around 50. LM-KT and EERNN showed lower performance in AUC and CSAUC than E2E or GRAM(1S), respectively.

As shown in Figure \ref{fig:crcf}, we empirically confirm that standard E2E training converged to better local minima than CRCF, at all regularization hyper-parameter values used in \cite{coevoulutionary}. GRAM 1S even converges faster than all variants of CRCF while maintaining E2E performance. CRCF also showed larger degradation in CSAUC. We noticed that as the degree of regularization increases, time for convergence increases notably. 

\noindent \textbf{Speed Comparison: }
We first highlight the reduction in training time via GRAM. Across 4 datasets, GRAM 1S achieves \textbf{4.3$\times$} speedup of the E2E baseline, while GRAM 1E achieves acceleration of \textbf{43$\times$}. We observe GRAM 1E achieves most significant training time acceleration as expected, since the boost ratio $\mathcal{R}$ of Eq.\eqref{eq:boostratio} is the largest in epoch-wise alternation. The power of GRAM 1E multiplies as the size of dataset increases, achieving remarkable \textbf{146$\times$} speed up for the largest dataset, TOEIC. Among all datasets, TOEIC has the largest boost ratio $\mathcal{R}$ based on Table \ref{tab:datameta} which explains the largest efficiency gain. 

GRAM 1E also out-speeds EERNN at all datasets despite the fact that EERNN uses W2V embeddings and a single LSTM layer for its CE. In terms of GRAM's alternating period, we note that the impact on speed boost is not monotonic on some datasets, potentially due to increased variance of optimization switching between the CE component and the CF component.

Based on GCP GPU cost for ondemand A100 (\$2.93/hr), training E2E model in TOEIC costs \textbf{\$3,161}. GRAM 1E drastically reduces the training cost to \textbf{\$21}. 
\\


\begin{table}[h]
\small
\centering
\begin{tabular}{lp{0.5cm}p{0.5cm}p{0.6cm}p{0.6cm}p{1.2cm}}
\toprule
\multicolumn{1}{l|}{}           & AUC  & MRR  & nDCG & nDCG & Speed-up  \\
\multicolumn{1}{l|}{Method}           &      &      & @5   & @10  &           \\ \midrule
\multicolumn{6}{c}{(Title-only)}                                        \\ \midrule
\multicolumn{1}{l|}{E2E}        & 68.9 & 33.3 & 36.8 & 43.2 & 1(10.4hr) \\
\multicolumn{1}{l|}{GRAM 1S}    & \underline{69}   & 33.5 & 37.1 & \underline{43.4} & 2.5       \\
\multicolumn{1}{l|}{GRAM 10S}   & 68.6 & \underline{33.7} & \underline{37.3} & \underline{43.4} & 1.9       \\
\multicolumn{1}{l|}{GRAM 0.5E}  & 68.7 & 32.9 & 36.2 & 42.7 & 13.5      \\
\multicolumn{1}{l|}{GRAM 1E}    & 68.7 & 33.1 & 36.6 & 42.8 & \underline{17.3}      \\
\multicolumn{1}{l|}{NRMS}       & 67.2 & 33.3 & 35.5 & 42   & 13.9      \\
\multicolumn{1}{l|}{SpeedyFeed} & 68.3 & 33.4 & 36.6 & 43   & 2.0       \\
\multicolumn{1}{l|}{NoFineTune} & 66.8 & 32.4 & 35.7 & 41.9 & 33.5      \\ \midrule
\multicolumn{6}{c}{(Title+Body)}                                        \\ \midrule
\multicolumn{1}{l|}{E2E}        &      &      &      &      & 1(*202hr) \\
\multicolumn{1}{l|}{GRAM 0.5E}  & \textbf{69.6} & 34   & 37.6 & \textbf{44}   & 45        \\
\multicolumn{1}{l|}{GRAM 1E}    & 69.3 & \textbf{34.1} & \textbf{37.8} & \textbf{44}   & \textbf{56}          \\
\bottomrule
\end{tabular}
\caption{\label{tab:mindperf}Prediction Performance and Training Speed on MIND Dataset. Training time is reported for E2E training. Overall best results among content finetuning methods are marked in bold, and best results utilizing title only are underlined.}
\normalsize
\end{table}

\subsection{News Recommendation}

\noindent \textbf{Performance Comparison: }
As shown in Table \ref{tab:mindperf}, Single-step GRAM matches performance of E2E training, and Multi-step GRAM shows less than 0.5\% performance loss. Multi-step GRAM's capability to incorporate abstract and body of the news article (Title+Body) significantly improved the performance beyond all methods relying on title information alone. 

SpeedyFeed \cite{speedyfeed} shows worse performance than all GRAM methods. This may due to SpeedyFeed's cache mechanism, as it fails to optimize news representations that were generated in recent time steps, unlike GRAM. We also noticed that SpeedyFeed's performance is highly sensitive to its hyper-parameters on the method's cache policy. Increasing max cache step hyper-parameter for faster training easily caused the training loss to spike, deteriorating the model convergence.

An ensemble of Single-step GRAM and Multi-step GRAM is currently ranked 4th in the MIND official leaderboard provided by Microsoft.\footnote{https://msnews.github.io/} Even without state-of-the-art CF module (Fastformer \cite{fastformer}) and CE module (UniLM \cite{unilm}), the ability to encode the body of the news article with GRAM shows a comparable performance with state-of-the-art News Recommendation models. 

\noindent \textbf{Speed Comparison: }
Multi-step GRAM shows consistent speed boost in MIND dataset, where Title-only GRAM 1E is \textbf{17.3$\times$} faster than E2E. SpeedyFeed's acceleration, on the other hand, was lower than that of GRAM 1S. Although SpeedyFeed's central batching collects unique items in the forward pass, it still requires gradient computations for PLM's weights on every interaction during backpropagation, having limited speed gain. Also, SpeedyFeed's auto-regressive formulation, the most significant factor of speed boost, was not applicable to MIND dataset, which does not provide negative samples (news impression list) per each positive interaction (news click) step. 

While \citet{nrms} reported that using all textual information increases model performance significantly in a small version of MIND dataset, baseline results from \citet{nrms, nrmsplm} only utilize news title due to computational complexity. With eight A100 GPUs, E2E training with all textual features is estimated to require \textbf{202} hours \footnote{Estimated time is calculated based on per epoch time for E2E with all texts and number of epochs to converge for Title-only E2E.} (\textbf{\$4,735} of training cost) to converge. For this reason, we were unable to produce the result for E2E with all textual information. In contrast, GRAM 1E requires only \textbf{3.6} hrs (\textbf{\$84}) to converge, requiring \textbf{56$\times$} less training time compared with E2E. 


\begin{table}[h]
\small
\centering
\begin{tabular}{p{1.4cm}c|cc}
\toprule
        & CE Batch          & \multicolumn{2}{c}{GPU Memory in \%, (Gb)} \\
Method  & Size & TOEIC                  & MIND                  \\
\midrule
E2E     &      N/A         & 95.2, (38.6)         & 95.1, (38.4)        \\
GRAM 1E & 8             & 12.1, (4.8)          & 12.5, (5.1)         \\
       & 32            & 16.0, (6.5)          & 34.9, (14.1)       \\
\bottomrule
\end{tabular}
\caption{\label{tab:utilgpu}GPU Memory Consumption, with CF batch size of 4 in single NVIDIA A100. E2E doesn't have CE batch size as CE module naturally receives all items included in CF batch size as input.}
\normalsize
\end{table}

\subsection{GPU Memory Consumption}
In E2E training, the entire computational graph as well as activations of all layers should be stored, resulting in a large GPU memory footprint \cite{e2ealternative}. In this perspective, splitting CF module and CE module in GRAM brings down the required GPU memory during computation. As (1) the model size of the CF module is relatively small (single LSTM/MHSA) and (2) CE module of Single-step GRAM updates all item representations in a given batch in one step, the memory reduction is not significant in Single-step GRAM. 

However, Multi-step GRAM can bring down the memory requirement significantly as its CE module uses multiple steps to update item representations. Table \ref{tab:utilgpu} compares the memory consumption between E2E and GRAM 1E with content encoder batch size of 8 and 32, respectively. Overall, GRAM 1E consumes less than 40\% of memory compared to E2E. While only GRAM 1E is compared, Multi-step GRAM in general consumes similar memory given the same content encoder batch size.

\section{Conclusion and Future Work}
In this paper, we proposed GRAM as an efficient method to train content-based collaborative filtering models. 
Single-step GRAM splits the CE module and CF module during training, accumulating gradients for items appearing repeatedly in a batch. This effectively reduces the number of CE module's gradient computation and negates the need to store the intermediate activations for both of the modules at once. Extending Single-step GRAM, we increase the gradient accumulation latency for Multi-step GRAM, gaining additional training speed boost and memory footprint reduction.

GRAM was empirically evaluated on 5 different tasks to demonstrate its efficiency and comparable prediction power. Utilizing GRAM's efficiency, Knowledge Tracing model trained with GRAM 1E will be deployed in Santa\footnote{https://www.aitutorsanta.com/}, an AI-powered English learning platform with 4 million users. 

{A limitation of our method is that we introduce an additional hyperparameter of gradient accumulation latency for Multi-step GRAM. We expect a more sophisticated gradient accumulation scheme may adaptively choose the gradient accumulation latency. Another potential extension of our research is to scale up GRAM for CCF tasks involving higher-dimensional inputs, such as images and videos.}



\section*{Acknowledgements}

We thank the anonymous reviewers of ACL Rolling Review, Seunghyun Lee (Riiid AI Research), and Suyeong An (Riiid AI Research) for their helpful feedback. We also thank Wansoo Kim (Riiid Infra) for helping us scale up our experiments to the Google Cloud Platform. 




\bibliography{anthology,custom}
\bibliographystyle{acl_natbib}
\clearpage
\appendix
\section{Proposition}





\label{appendix:proof}
\begin{proof}
\small
\begin{align}
    \frac{\partial L}{\partial \theta_f} &= \frac{\partial L}{\partial h} 
    \frac{\partial h}{\partial \theta_f} = \frac{\partial L}{\partial y} 
    \frac{\partial y}{\partial h} \frac{\partial h}{\partial \theta_f}\\
    \frac{\partial \tilde{L}}{\partial \theta_f'}\bigg|_{\theta_f'=\theta_f} &= 
    -\frac{\partial f}{\partial \theta_f'} \cdot
    \bigg(f(x;\theta_f) - \frac{\partial L}{\partial y}
    \frac{\partial y}{\partial h} - f(x;\theta_f') \bigg)\\
    &= \frac{\partial L}{\partial y} 
    \frac{\partial y}{\partial h} \frac{\partial h}{\partial \theta_f}
\end{align}
\normalsize
\end{proof}

\section{Hierarchical illustration of GRAM}
\label{appendix:illustration}
\begin{figure}[h]
    \centering
    \includegraphics[width=0.8\columnwidth]{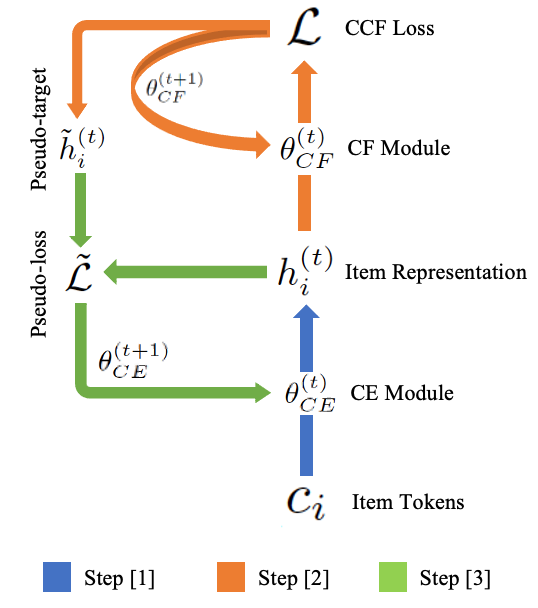}
    \caption{Gradient on item representations are accumulated on Pseudo-target to mimic E2E training.}
    \label{fig:algoscheme}
\end{figure}

Figure \ref{fig:algoscheme} is the hierarchical illustration of GRAM. Refer to Section \ref{sec4:gramalgo} for more details.

\begin{figure*}[h]
    \centering
    \includegraphics[width=1.9\columnwidth]{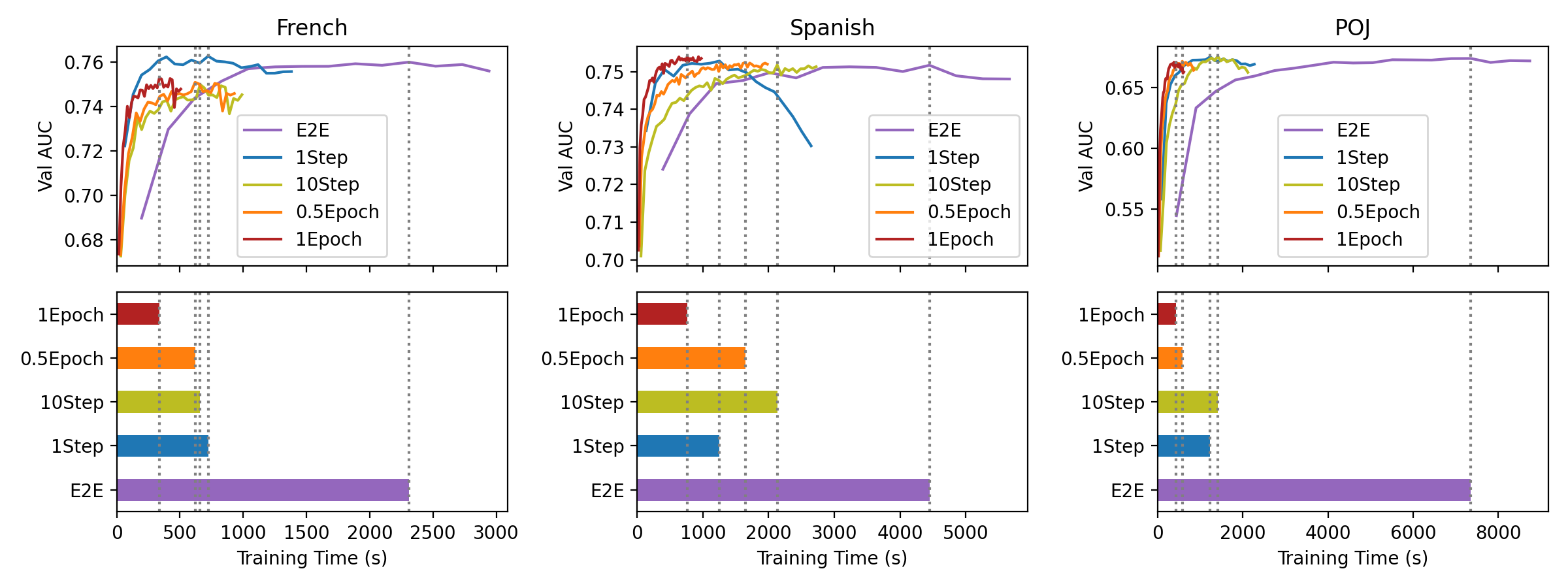}
    \caption{\label{fig:comparison}. Validation AUC convergence and training time until early-stopped checkpoint, across different GRAM alternation periods. Color schemes are synchronized across all subplots.}
\end{figure*}

\section{Validation Performance}
Figure \ref{fig:comparison} shows the validation performance per training time for E2E and GRAM variants. As the graph includes 10 epochs for early stopping patience, some overfitting is witness at the end of the training curve, such as the sudden drop of GRAM 1S in Spanish dataset after marking best performance.

\section{Experiment Detail}
\label{appendix:detail}
All experiments are ran 3 times and averaged results are reported except TOEIC E2E and MIND E2E, due to high computational cost mentioned in the result section. Across all experiments, CE component uses TinyBERT\cite{tinybert}, a distilled BERT with on-par performance. The architecture contains 6 MHSA layers of dimension 768. CF component for KT uses 2-layer LSTM network, following \citet{dkt}. In NR, we use a single MHSA layer, following \citet{nrms, nrmsplm}.

In both domains, learning rate of 1e-4 was used for CF module and CE module after learning rate ablation in the scope of [1e-3, 1e-4, 1e-5]. Adam with Noam scheduler was used as the optimizer. Test metrics were measured by the best validation checkpoint after early stopping of patience 10 epochs. As E2E training consumes different amount of memory based on item token length, different batch sizes were used across datasets. Batch sizes per datasets are the following: 32 for Duolingo, POJ, and TOEIC, and 256 for MIND. As Multi-step GRAM requires much less memory compared with E2E, higher batch size was able to be used for large datasets such as TOEIC and MIND. Details are mentioned in Section C. 

For NoFinetune experiment, pre-computed item representations from CE were initialized to CF's item embeddings to boost training speed.

Authors of EERNN also proposed EKT\cite{ekt}, which explicitly models the student's knowledge state for different knowledge concepts. As knowledge concept labels are not available in most datasets, we do not test EKT.

\subsection{Knowledge Tracing}
Mean-pooling was used to extract question representation from contextualized token embedding. For relatively small KT datasets (Duolingo and POJ), it was challenging to secure meaningful number of cold-start items (questions) in the test split. For these datasets, additional cold-start questions were randomly picked and interactions on those questions in the training split were removed to secure meaningful number of cold start interactions. Items with token length over max seq len (512) were truncated. For TOEIC dataset, passage, question, and choices were concatenated as content token sequence for CE component.
\subsection{News Recommendation}
For NR, we follow \cite{nrmsplm} to use additive attention based pooling to extract news article representation. Title, abstract, body were concatenated for Setting B, with max len of 24 for title, 50 for abstract, and 400 for body. Items with token length over max seq len were truncated.

\section{Result Detail}
\label{appendix:result_detail}
\subsection{E2E vs Single-step GRAM Confidence Interval}
\begin{table}[h]
\small
\begin{tabular}{l|ll}
\toprule
Metrics           & E2E                               & GRAM (1S)      \\
\midrule
French Total AUC  & 0.748 (0.001)                     & 0.750 (0.0006) \\
Spanish Total AUC & \multicolumn{1}{c}{0.751(0.0005)} & 0.751(0.0006)  \\
POJ Total AUC     & 0.654 (0.0009)                    & 0.655 (0.0011) \\
\bottomrule
\end{tabular}
\normalsize
\caption{\label{tab:ci}}
\end{table}
As TOEIC and MIND E2E result is from a single run due to high computational complexity, Test AUCs on remaining three datasets (Duolingo French, Spanish, POJ) are reported with 95\% CIs in Table \ref{tab:ci}

\section{Dataset Collection Methods}
\label{appendix:dataset_collection}

\textbf{Duolingo}: gathered from 2018 Duolingo Shared Task on Second Language Acqui-sition Modeling.

\noindent\textbf{POJ}: publicly available question texts and interaction logs were scraped from their public \href{http://poj.org/}{website}.

\noindent\textbf{TOEIC}: content materials for corresponding question IDs in the dataset were collected privately. 

\noindent\textbf{MIND}: collected from \href{https://msnews.github.io/}{website}

\noindent
Duolingo (French, Spanish), POJ, MIND datasets are free to download for research purposes under respective terms. Interaction data for TOEIC is avilable as well for research purposes.

\end{document}